\documentclass[10pt,twocolumn,letterpaper]{article}

\usepackage{cvpr}
\usepackage{times}
\usepackage{balance}
\usepackage{graphicx}
\usepackage{amsmath,amssymb}
\usepackage[pagebackref=false,breaklinks=true,colorlinks,bookmarks=false,urlcolor=blue,linkcolor=blue,citecolor=blue,pdftitle={MRF-based Background Initialisation for Improved Foreground Detection in Cluttered Surveillance Videos},pdfauthor={Vikas Reddy, Conrad Sanderson, Andres Sanin, Brian C. Lovell}]{hyperref}

\cvprfinalcopy

\def\cvprPaperID{xx}

\begin{document}

\title
  {
  MRF-based Background Initialisation for Improved Foreground Detection\\
  in Cluttered Surveillance Videos
  }

\author
  {
  Vikas Reddy,
  Conrad Sanderson,
  Andres Sanin,
  Brian C. Lovell
  \\
  ~\\
  NICTA, PO Box 6020, St Lucia, QLD 4067, Australia \thanks{{\bf Published~in:} Lecture Notes in Computer Science (LNCS), Vol.~6494, pp.~547-559, 2011. \href{http://dx.doi.org/10.1007/978-3-642-19318-7_43}{http://dx.doi.org/10.1007/978-3-642-19318-7\_43}}
  \\
  The University of Queensland, School of ITEE, QLD 4072, Australia
  }

\maketitle
\thispagestyle{empty}


\begin{abstract}

\noindent
Robust foreground object segmentation via background
modelling is a difficult problem in cluttered environments,
where obtaining a clear view of the background to model is almost impossible.
In this paper, we propose a method capable of robustly estimating the background
and detecting regions of interest in such environments.
In particular, we propose to extend the background initialisation component
of a recent patch-based foreground detection algorithm
with an elaborate technique based on Markov Random Fields,
where the optimal labelling solution is computed using iterated conditional modes.
Rather than relying purely on local temporal statistics,
the proposed technique takes into account the spatial continuity of the entire background.
Experiments with several tracking algorithms on the CAVIAR dataset indicate that
the proposed method leads to considerable improvements in object tracking accuracy,
when compared to methods based on Gaussian mixture models and feature histograms.

\end{abstract}

\section{Introduction}

One of the low-level tasks in most intelligent video surveillance applications
(such as person tracking and identification)
is to segment objects of interest from an image sequence.
Typical segmentation approaches employ the idea of comparing each frame against a model of the background,
followed by selecting the outliers (i.e.,~pixels or areas that do not fit the model).
However, most methods presume the training image sequence used to model the background is free from foreground objects.
This assumption is often not true in the case of uncontrolled environments such as train stations and motorways,
where directly obtaining a clear background is almost impossible.
Furthermore, in outdoor video surveillance a strong illumination change can render the existing background model ineffective
(e.g.,~due to introduction of shadows~\cite{Sanin_PR_inpress}),
thereby forcing us to compute a new background model.
In such circumstances,
it becomes inevitable to reinitialise the background model using cluttered sequences 
(i.e.,~where parts of the background are occluded).
Robust background initialisation in these scenarios can result in improved segmentation of foreground objects,
which in turn can lead to more accurate tracking.

The majority of the algorithms described in the literature, 
such as~\cite{kim2005rtf,maddalena2008self,Reddy_TCSVT_inpress,toyama1999wpa},
do~not have a robust strategy to handle cluttered sequences.
Specifically, they fail when the background in the training sequence
is exposed for a shorter duration than foreground objects.
This is due to the model being initialised by relying solely on the temporal statistics of the image data, which is
easily affected by the inclusion of foreground objects in the training sequence.

To alleviate this problem, a few algorithms have been proposed to initialise the background image
from cluttered image sequences.
Typical examples include median filtering, finding pixel intervals of 
stable intensity in the image sequence~\cite{wang2006nrs},
building a codebook for the background model~\cite{kim2005rtf},
agglomerative clustering~\cite{colombari2006bic}
and minimising an energy function using an \textit{$\alpha$--expansion} algorithm~\cite{Boykov_1999}.
However, none of them evaluate the foreground segmentation accuracy
using their estimated background model.

In this paper, we propose to replace the background model initialisation component
of a recently introduced foreground segmentation method~\cite{Reddy_TCSVT_inpress}
and show that the performance can be considerably improved in cluttered environments.
The proposed background initialisation is carried out in a Markov Random Field (MRF) framework,
where the optimal labelling solution is computed using iterated conditional modes.
The spatial continuity of the background is also considered in addition to the temporal statistics of the training sequence.
This strategy is particularly robust to training sequences containing foreground objects
exposed for longer duration than the background over a given time interval.

Experiments on the CAVIAR dataset, where most of the sequences contain occluded backgrounds, 
show that the proposed framework  (MRF {\footnotesize{+}} multi-stage classifier) yields considerably better 
results in terms of tracking accuracy than the baseline multi-stage classifier method~\cite{Reddy_TCSVT_inpress}
as well as methods based on Gaussian mixture models~\cite{stauffer1999abm}
and feature histograms~\cite{li2003foreground}.

We continue as follows.
The overall foreground segmentation framework is described in~Section~\ref{sec:Background_Subtraction_Framework},
followed by the details of the proposed MRF-based background initialisation method in Section~\ref{subsec:Markov_Random_Field_approach}.
Performance evaluations and comparisons
with three other algorithms are given in Section~\ref{sec:Experiment_Results},
followed by the main findings in Section~\ref{sec:Conclusion}.


\section{Foreground Segmentation Framework}
\label{sec:Background_Subtraction_Framework}

We build on the patch-based multi-stage foreground segmentation method proposed in~\cite{Reddy_TCSVT_inpress},
which has four major components:
\begin{enumerate}

\item
Division of a given image into overlapping blocks (patches),
followed by generating a low-dimensional 2D Discrete Cosine Transform~(DCT) based descriptor for each block~\cite{Gonzales_2007}.

\item
Classification of each block into foreground or background based on a background model,
where each block is sequentially processed by up to three classifiers.
As soon as one of the classifiers deems that the block is part of the background,
the remaining classifiers are not consulted.
In sequential order of processing, the three classifiers are:

\begin{enumerate}
\renewcommand{\labelenumii}{(\alph{enumii})}

\item
a probability measurement according to a location specific multivariate Gaussian model of the background
(i.e.,~one Gaussian for each block location);

\item
an illumination robust similarity measurement through a cosine distance metric;

\item
a temporal correlation check where
blocks and decisions from the previous image are taken into account.

\end{enumerate}

\item
Model reinitialisation to address scenarios where a sudden and significant scene change
can make the current model inaccurate.

\item
Probabilistic generation of the foreground mask,
where the classification decisions for all blocks are integrated.
The overlapping nature of the analysis is exploited to produce smooth contours
and to minimise the number of errors (both false positives and false negatives).

\end{enumerate}

\noindent
Parts 2(a) and 2(b) require a location specific Gaussian model,
which can be characterised by a mean vector $\boldsymbol{\mu}$ and covariance matrix $\boldsymbol{\Sigma}$.
In an attempt to allow the training sequence to contain moving foreground objects,
a rudimentary Gaussian selection strategy is employed in~\cite{Reddy_TCSVT_inpress}.
Specifically, for each block location a two-component Gaussian mixture model~(GMM) is trained,
followed by taking the absolute difference of the weights of the two Gaussians.
If the difference is greater than $0.5$,
the Gaussian with the dominant weight is retained.
The reasoning is that the less prominent Gaussian
is modelling moving foreground objects and/or other outliers.
If the difference is less than $0.5$,
it is assumed that no foreground objects are present
and all available data for that particular block location is used to estimate the parameters of the single Gaussian.

There are several problems with the above parameter selection approach.
It is assumed that foreground objects are either continuously moving in the sequence
or that no object stays in one location for more than 25\% of the length of the training sequence.
This is not guaranteed to occur in uncontrolled environments such as railway stations.
The decision to retain the dominant Gaussian solely relies on local temporal statistics
and ignores rich local spatial correlations that naturally exist within a scene.

To address the above problems, we propose to estimate the parameters of the background model 
via a Markov Random Field~(MRF) framework,
where in addition to temporal information,
spatial continuity of the entire background is considered.
The details of the MRF-based algorithm are given in the following section.


\section{Proposed Background Initialisation}
\label{subsec:Markov_Random_Field_approach}

Let the resolution of the image sequence {$I$} be {$\mathcal{W} \times \mathcal{H}$}, with $\phi$ colour channels.
The proposed algorithm has three main stages:
{\bf (1)}~division of each frame into non-overlapping blocks and collection of possible background blocks over a given time interval,
{\bf (2)}~partial background reconstruction using unambiguous blocks,
{\bf (3)}~ambiguity resolution through exploitation of spatial correlations across neighbouring blocks.
An example of the algorithm in action is shown in Fig.~\ref{fig:BGI_iterations}.
The details of the three stages are given below.

In stage~1, each frame is viewed as an instance of an undirected graph,
where the nodes of the graph are blocks of size {$N \times N \times \phi$} pixels%
\footnote
  {
  For implementation purposes, each block location and its instances at every frame are treated as a node and its labels, respectively.
  }%
.
We denote the nodes of the graph by {$\mathcal{N}{(i,j)}$}
for \mbox{$i = 0,1,2,\cdots,(\mathcal{W}/N) - 1$},  $j = 0,1,2,\cdots,(\mathcal{H}/N) - 1$.
Let {$I_f$} be the \mbox{$f$-th} frame of the training image sequence
and let its corresponding node labels be denoted by {$\mathcal{L}_f(i,j)$},
and {$f = 1,2,\cdots,F$},
where $F$ is the total number of frames.
For convenience, each node label {$\mathcal{L}_f(i,j)$} is vectorised
into an {${\phi}N^2$} dimensional vector {$\textbf{l}_f(i,j)$}. 
In comparison to pixel-based processing, block-based processing 
is more robust against noise and captures better contextual spatial continuity of the background.

At each node {$(i,j)$},
a~representative set {$\mathcal{R}(i,j)$} is maintained.
It contains only unique representative labels,
{${\textbf{r}}_k(i,j)$} for {$k = 1, 2, \cdots, S$} (with {$S \leq F$})
that were obtained along its temporal line.
To determine uniqueness, the similarity of labels is calculated as described in Section~\ref{subsec:Similarity_criteria_for_labels}.
Let weight {$W_{k}$} denote the number of occurrences of {${\textbf{r}}_k$} in the sequence,
i.e.,~the number of labels at location $(i,j)$ which are deemed to be the same as {${\textbf{r}}_k(i,j)$}.

It is assumed that one element of {$\mathcal{R}(i,j)$} corresponds to the background.
To ensure labels corresponding to moving objects are not stored,
label ${\textbf{b}}_f(i,j)$ will be registered as ${\textbf{r}}_{k+1}(i,j)$
only if it appears in at least $f_{min}$ consecutive frames,
where $f_{min}$ ranges from $2$ to $5$.

In stage~2, representative sets {$\mathcal{R}(i,j)$}  having just one label
are used to initialise the corresponding node locations {$\mathcal{B}(i,j)$}
in the background {$\mathcal{B}$}.

In stage~3, the remainder of the background is estimated iteratively.
An optimal labelling solution is calculated by considering the likelihood of each of its labels
along with the \textit{a~priori} knowledge of the local spatial neighbourhood modelled as an MRF.
Iterated conditional mode (ICM), a deterministic relaxation technique, performs the optimisation.

The MRF framework is described in Section~\ref{subsec:MAPMRF_Framework}.
The strategy for selecting the location of an empty background node to initialise a label
is described in Section~\ref{subsec:Neighbourhood_selection}.
The procedure for calculating the energy potentials, a prerequisite in determining the \textit{a~priori} probability,
is described in Section~\ref{subsec:Calculation_of_energy_potential}.
In Section~\ref{subsec:Modified_Background_Model},
the background model (used by the foreground segmentation algorithm overviewed in Section~\ref{sec:Background_Subtraction_Framework})
is modified using the estimated background frame.

\begin{figure*}[!t]
  \begin{minipage}{1.0\textwidth}
    \includegraphics[width=0.2433\columnwidth]{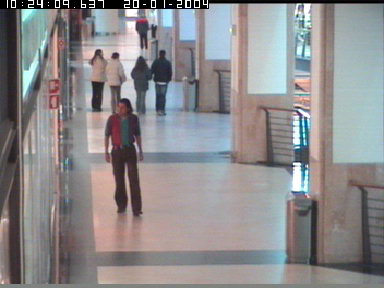}
    \hfill
    \includegraphics[width=0.243\columnwidth]{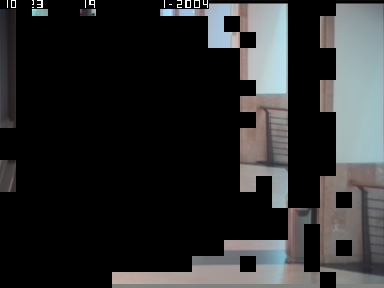}
    \hfill
    \includegraphics[width=0.243\columnwidth]{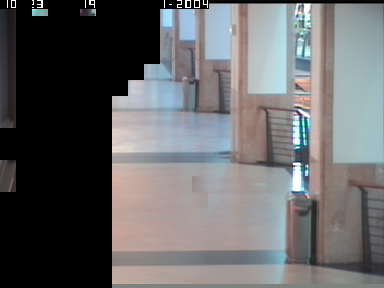}
    \hfill
    \includegraphics[width=0.243\columnwidth]{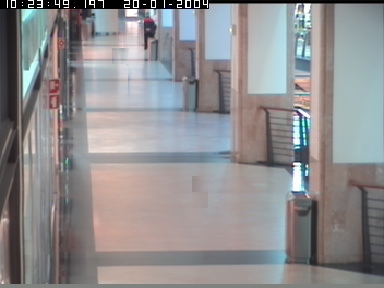}
  \end{minipage}

  ~

  \begin{minipage}{1.0\textwidth}
    \begin{minipage}{0.243\columnwidth}
      \centerline{\bf(i)}
    \end{minipage}
    \hfill
    \begin{minipage}{0.243\columnwidth}
      \centerline{\bf(ii)}
    \end{minipage}
    \hfill
    \begin{minipage}{0.243\columnwidth}
      \centerline{\bf(iii)}
    \end{minipage}
    \hfill
    \begin{minipage}{0.243\columnwidth}
      \centerline{\bf(iv)}
    \end{minipage}
  \end{minipage}
  \caption
    {
    Example of background estimation from an image sequence cluttered with foreground objects:
    {\bf (i)}~example frame,
    {\bf (ii)}~partial background initialisation (after stage~2),
    {\bf (iii)}~remaining background estimation in progress (stage~3),
    {\bf (iv)}~estimated background.
    }
  \label{fig:BGI_iterations}
\end{figure*}

\subsection{Similarity Criteria for Labels}
\label{subsec:Similarity_criteria_for_labels}

Two labels ${\textbf{l}}_f(i,j)$ and ${\textbf{r}}_k(i,j)$ are similar
if the following two constraints are satisfied:
\begin{equation}
  \frac
  { \left( {\textbf{r}}_{k}(i,j) - { \mu_{r_k}}(i,j) \right)' \left( {\textbf{l}}_{f}(i,j) - { \mu_{b_f}}(i,j) \right)}
  { {\sigma_{r_k} \sigma_{b_t}}}
  ~>~ \mathcal{T}_1
  \label{eqn:corr_ptrk}
\end{equation}

\noindent
and
\begin{equation}
  \frac{1}{\phi N^2} \sum\nolimits_{n = 0}^{\phi N^2 - 1} \left|{\textit{d}}_{k_n}(i,j)\right|    ~<~ \mathcal{T}_2
  \label{eqn:diff_ptrk}
\end{equation}

\noindent
where $\mu_{r_k},\mu_{l_f}$ and $\sigma_{r_k},\sigma_{l_f}$
are the mean and standard deviation of the elements of labels $\textbf{r}_{k}$ and $\textbf{l}_f$ respectively,
while {${\textbf{d}}_k(i,j) = {\textbf{l}}_f(i,j) - {\textbf{r}}_k(i,j)$}.

Eqns.~(\ref{eqn:corr_ptrk}) and (\ref{eqn:diff_ptrk})
respectively
evaluate the correlation coefficient and the mean of absolute differences (MAD) between the two labels.
The former constraint ensures that labels have similar texture/pattern while the latter one ensures that
they are close in {$\phi N^2$} dimensional space.
In contrast, we note that in~\cite{colombari2006bic} the similarity criteria
is based just on the sum of squared distances between the two blocks.

{$\mathcal{T}_1$} is selected empirically (typically 0.8),
to ensure that two visually identical labels are not treated as being different due to image noise.
{$\mathcal{T}_2$}~is proportional to image noise.

\subsection{Markov Random Field (MRF) Framework}
\label{subsec:MAPMRF_Framework}

MRF has been widely employed in solving problems in image processing that can
be formulated as labelling problems~\cite{besag1986sad,sheikh2005bayesian}.

Let {$\textbf{X}$} be a 2D random field,
where each random variate {$\textit{X}_{(i,j)}$ $( \forall \hspace{2pt}i,j )$}
takes values in discrete \textit{state space} {$\Lambda$}.
Let {$\omega$} {$\in$} {$\Omega$} be a \textit{configuration}
of the variates in {$\textbf{X}$}, and let {$\Omega$} be the set of all such configurations.
The joint probability distribution of  $\textbf{X}$ is considered Markov if
\begin{equation}
 {  p(\textbf{X} = \omega) > 0, \hspace{2pt} \forall \hspace{3pt}\omega \in \Omega }
   \label{eqn:mrf_conds1}
\end{equation}

\noindent
and
\begin{equation}
    p\left(\textit{X}_{(i,j)} | \textit{X}_{(a,b)}, {\textsl{\footnotesize{(i, j)}} \neq \textsl{\footnotesize{(a, b)}}}\right) \\
    = p\left(\textit{X}_{(i,j)} | \textit{X}_{\mathcal{N}_{(i,j)}}\right)
\label{eqn:mrf_conds1b}
\end{equation}

\noindent
where {$\textit{X}_{\mathcal{N}_{(i,j)}}$} refers to the local \textit{neighbourhood system} of {$\textit{X}_{(i,j)}$}.

Unfortunately, the theoretical factorisation of the joint probability distribution of the MRF
turns out to be intractable.
To simplify and provide computationally efficient factorisation,
Hammersley-Clifford theorem~\cite{besag1974sia}
states that an MRF can equivalently be characterised by a Gibbs distribution.
Thus
\begin{equation}
    p(\textbf{X}= \omega) = {e^{-U(\omega)/T}}~\textbf~/~\left({\sum\nolimits_{\omega}{e^{-U(\omega)/T}}}\right)
\label{eqn:mrf_defines1}
\end{equation}

\noindent
where the denominator is a normalisation constant known as the \textit{partition function},
{$T$} is a constant used  to moderate the peaks of the distribution
and {$U(\omega)$} is an \textit{energy function}
which is the sum of \textit{clique/energy potentials} {$V_c$} over all possible cliques~$C$:
\begin{equation}
    U(\omega) = \sum\nolimits_{c \in \textit{C}}V_c(\omega)
    \label{eqn:mrf_defines3}
\end{equation}

\noindent
The value of {$V_c(\omega)$} depends on the local configuration of clique {$c$}.

In our framework, information from two disparate sources is combined using Bayes' rule.
The local visual observations at each node to be labelled yield label likelihoods.
The resulting label likelihoods are combined with \textit{a~priori} spatial knowledge of the neighbourhood represented as an MRF.

Let each input image {$I_f$} be treated as a realisation of the random field {$\mathcal{B}$}.
For each node {$\mathcal{B}(i,j)$}, the  representative set {$\mathcal{R}(i,j)$}
containing unique labels is treated as its \textit{state space} with each {$\textbf{r}_k(i,j)$} as its plausible label%
\footnote{To simplify the notations, index term $(i,j)$ has been omitted from here onwards.}%
.

Using Bayes' rule, the posterior probability for every label at each node is derived
from the \textit{a~priori} probabilities and the observation-dependent likelihoods given by:
\begin{equation}
   P(\textbf{r}_{k}) = l(\textbf{r}_{k})p(\textbf{r}_{k})
\label{eqn:bayestheorem1}
\end{equation}

\noindent
The product is comprised of  likelihood {$l(\textbf{r}_k)$} of each label
{$\textbf{r}_{k}$} of set {$\mathcal{R}$} and its \textit{a~priori} probability density
{$ p(\textbf{r}_{k})$}, conditioned on its local neighbourhood.
In the derivation of likelihood function it is assumed that at each node the observation components {$\textbf{r}_k$}
are conditionally independent and have the same known conditional density function dependent only on that node.
At a given node, the label that yields maximum \textit{a posteriori} (MAP) probability
is chosen as the best continuation of the background at that node.

To optimise the MRF-based function defined in Eqn.~(\ref{eqn:bayestheorem1}),
ICM is used since it is computationally efficient and avoids large scale effects%
\footnote{An undesired characteristic where a single label is wrongly assigned to most of the~nodes of the random field.}%
~\cite{besag1986sad}.
ICM  maximises local conditional probabilities iteratively until convergence is achieved.
In ICM an initial estimate of the labels is typically obtained by maximising the likelihood function.
However, in our framework an initial estimate consists of partial reconstruction of the background
at nodes having just one label which is assumed to be the background.
Using the available background information, the remaining unknown background is estimated progressively
(see Section~\ref{subsec:Neighbourhood_selection}).

At every node, the likelihood of each of its labels {$\textbf{r}_k$ $(k = 1, 2, \cdots, S)$}
is calculated using corresponding weights {$W_k$}. 
The higher the occurrences of a label, the more is its likelihood to be part of the background.
Empirically, the likelihood function is modelled by a simple weighted function,
given by:
\begin{equation}
 l(\textbf{r}_k)= {W_{c_k}}/{\sum\nolimits_{k=1}^{S}{W_{c_k}}}
    \label{eqn:rayleigh}
\end{equation}%

\noindent
where {$W_{c_k} = \operatorname{min}(W_{max},W_k)$}. Capping the weight 
is necessary in circumstances where the image sequence
has a stationary foreground object visible for an exceedingly long period.

The spatial neighbourhood modelled as Gibbs distribution
(Eqn.~(\ref{eqn:mrf_defines1})) is encoded into an \textit{a~priori} probability density.
The formulation of the clique potential {$V_c(\omega)$} referred in Eqn.~(\ref{eqn:mrf_defines3})
is described in the Section~\ref{subsec:Calculation_of_energy_potential}.
Using  Eqns.~(\ref{eqn:mrf_defines1})~and~(\ref{eqn:mrf_defines3})
the calculated clique potentials $V_c(\omega)$ are transformed into \textit{a~priori} probabilities.
For a given label, the smaller the value of energy function, the
greater is its probability in being the best match with respect to its neighbours.

In our evaluation of the posterior probability given by Eqn.~(\ref{eqn:bayestheorem1}), more emphasis
is given to the local spatial context term than the likelihood function which is based 
on mere temporal statistics.
Thus, taking log of Eqn.~(\ref{eqn:bayestheorem1}) and assigning a weight to the prior, we get:
\begin{equation}
  {\log\left(P(\textbf{r}_{k})\right) =
  \log \left(l(\textbf{r}_{k})\right) + \eta \hspace{1pt}
  \log \left(p(\textbf{r}_{k})\right)}
\label{eqn:bayestheoremlog}
\end{equation}

\noindent
where $\eta$ has been empirically set to number of neighbouring nodes used in clique potential calculation
(typically $\eta$ = 3).

\subsection{Node Initialisation}
\label{subsec:Neighbourhood_selection}

\begin{figure}[!b]
  \centering
  \includegraphics[width=0.5\columnwidth]{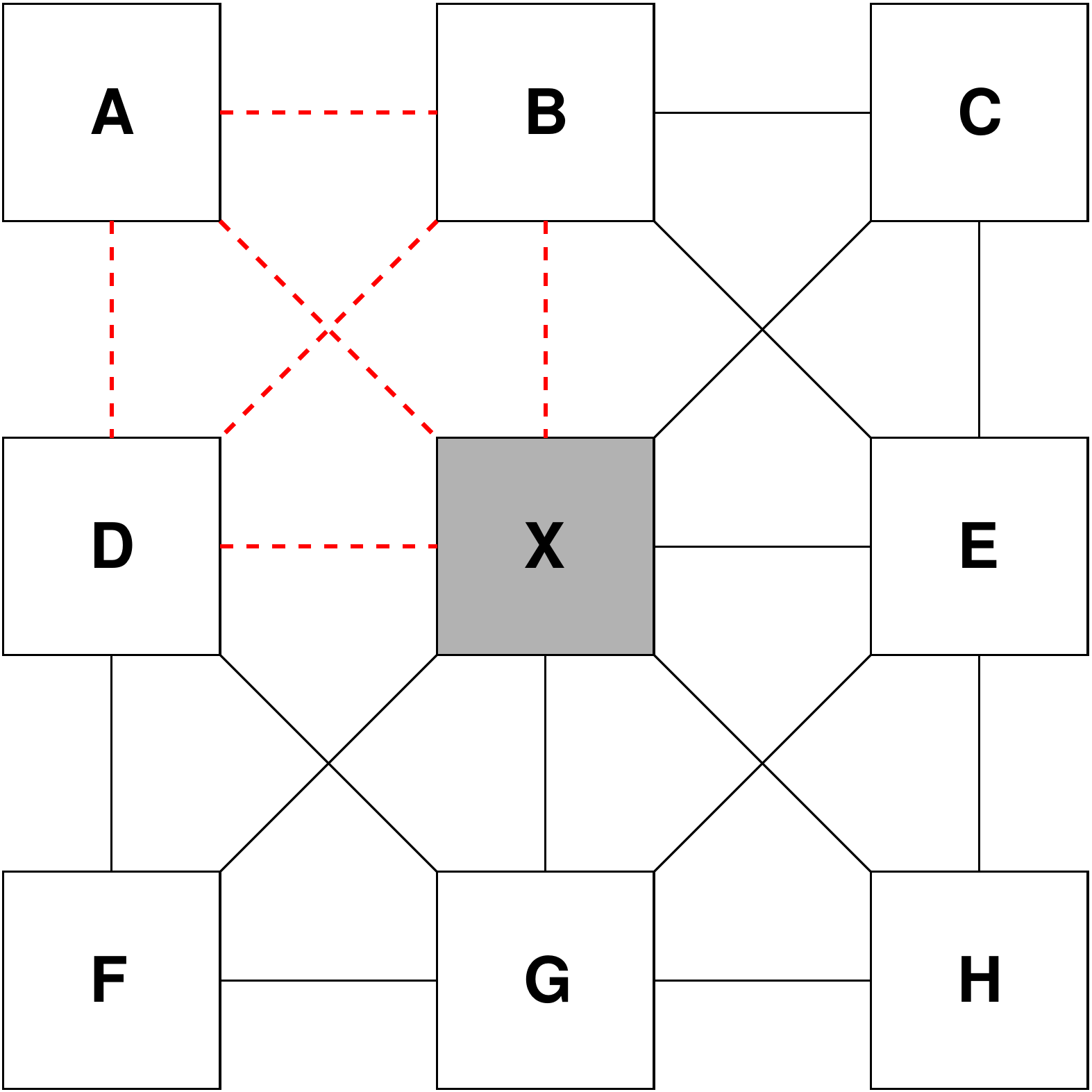}
  \caption
    {
    The local neighbourhood system and its four cliques.
    Each clique is comprised of 4 nodes (blocks).
    To demonstrate one of the cliques, the the top-left clique has dashed red links.
    }
  \label{fig:clique1}
\end{figure}

Nodes containing a single label in their representative set are directly initialised
with that label in the background (see Fig.~\ref{fig:BGI_iterations}(ii)).
However, in rare situations there's a possibility that all sets may contain more than 1 label 
(no trivial nodes). In such cases, the label having the largest weight from the representative 
sets of the 4 corner nodes is selected as an initial seed.
We assume at least 1 of the corner regions corresponds to a static region.
The rest of the nodes are initialised based on constraints as explained below.
In our framework, the local \textit{neighbourhood system}~\cite{geman1984srg} of a node and the corresponding cliques
are defined as shown in Fig.~\ref{fig:clique1}.
The background at an empty node will be assigned only if at least 2 neighbouring nodes of its \mbox{4-connected} neighbours
adjacent to each other and the diagonal node located between them are already assigned with background labels.
For instance, in Fig.~\ref{fig:clique1}, we can assign a label to node $X$ if at least nodes $B$, $D$ (adjacent 4-connected neighbours)
and $A$ (diagonal node) have already been assigned with labels.
In other words, label assignment at node $X$ is \textit{conditionally independent} of all other nodes given these 3 neighbouring nodes.

Let us assume that all nodes except $X$ are labelled.
To label node $X$ the procedure is as follows.
In Fig.~\ref{fig:clique1}, four cliques involving $X$ exist.
For each candidate label at node $X$, the energy potential for each of the four cliques
is evaluated independently given by Eqn.~(\ref{eqn:energypotential})
and summed together to obtain its energy value. The label that yields the least value is likely to be
assigned as the background.

Mandating that the background should be available in at least 3 neighbouring nodes
located in three different directions with respect to node $X$ ensures that the best match is obtained
after evaluating the continuity of the pixels in all possible orientations.

In cases where not all the three neighbours are available, 
to assign a label at node $X$ we use one of its 4-connected neighbours
whose node has already been assigned with a label.
Under these contexts, the clique is defined as two adjacent nodes either in the horizontal or vertical direction.

After initialising all the empty nodes an accurate estimate of the background is typically obtained.
Nonetheless, in certain circumstances an incorrect label assignment at a node may cause an error
to occur and propagate to its neighbourhood.
The problem is successfully redressed by the application of ICM.
In subsequent iterations, in order to avoid redundant calculations,
the label process is carried out only at nodes where a change in the label of one of their 8-connected neighbours
occurred in the previous iteration.

\subsection{Calculation of the Energy Potential}
\label{subsec:Calculation_of_energy_potential}

In Fig.~\ref{fig:clique1}, it is assumed that all nodes except {$X$} are assigned with the background labels.
The algorithm needs to assign an optimal label at node {$X$}.
Let node {$X$} have {$S$} labels in its state space {${\mathcal{R}}$} for {$k = 1, 2, \cdots, S$},
where one of them represents the true background.
Choosing the best label is accomplished by analysing the spectral response of every possible clique
constituting the unknown node $X$.
For the decomposition we chose the Discrete Cosine Transform (DCT)~\cite{Gonzales_2007}
in a similar manner to~\cite{Reddy_2009}.

We consider the top left clique consisting of nodes $A$, $B$, $D$ and {$X$}.
Nodes $A$, $B$ and $C$ are assigned with background labels. 
Node {$X$} is assigned with one of {$S$} candidate labels.
For each colour channel $z$, we take the 2D DCT of the resulting clique.
The transform coefficients are stored in matrix $\textbf{T}_{k}^{z}$ of size \mbox{$M \times M$} (\mbox{$M = 2N$})
with its elements referred to as {$T_k^{z}{(v,u)}$}.
The term {$T_{k}^{z}{(0,0)}$} (reflecting the sum of pixels at each node) is forced to 0
since we are interested in analysing the spatial variations of pixel values.

\begin{figure*}[!t]
\centering
    \begin{minipage}{1\textwidth}
      \begin{center}
  
   \begin{minipage}{0.10\textwidth}
          \begin{center}
            \begin{minipage}{0.85\textwidth}
              \begin{center}
                {\includegraphics[width=1\textwidth]{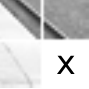}}
              \end{center}            
            \end{minipage}
    \end{center}

    ~

    ~
     \begin{minipage}{1\textwidth}
         \begin{center} 
            \begin{minipage}{1\textwidth}
              \begin{center}    
                {\footnotesize 1}~\includegraphics[width=0.4\textwidth]{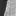}  
              \end{center}
            \end{minipage}

~

      \begin{minipage}{1\textwidth}
              \begin{center}    
                {\footnotesize 2}~\includegraphics[width=0.4\textwidth]{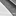} 
              \end{center}
            \end{minipage}
         \end{center}
  \end{minipage}
      \end{minipage}
        \hfill
        \vline
        \hfill
 \begin{minipage}{0.42\textwidth}
\begin{center}
            \begin{minipage}{0.20\textwidth}
              \centerline{\includegraphics[width=\textwidth]{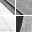}}
            \end{minipage}

            \begin{minipage}{1\textwidth}
              \centerline{\includegraphics[width=\textwidth]{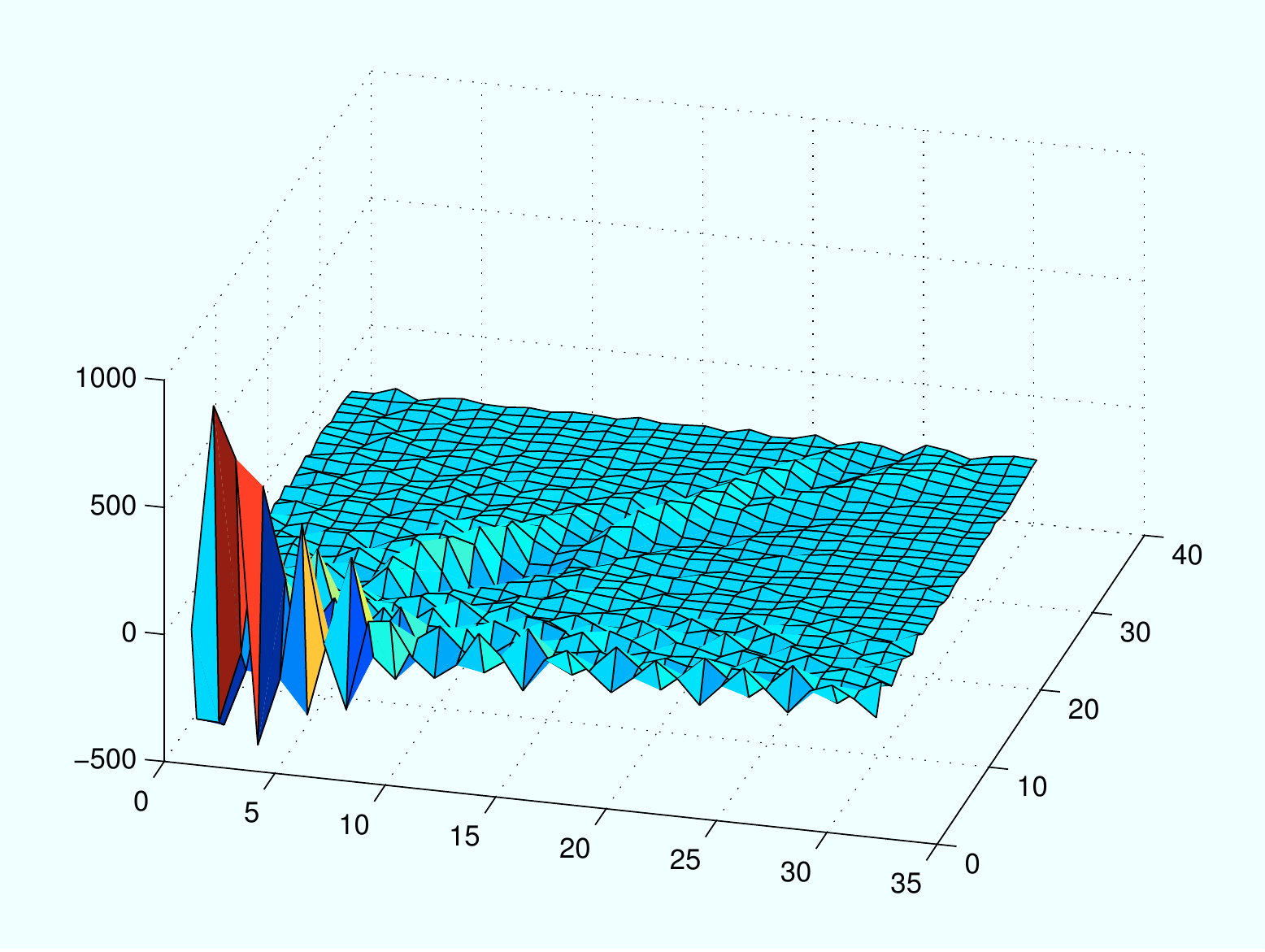}}
            \end{minipage}
\end{center}
        \end{minipage}
        \hfill
        \vline
        \hfill        
\begin{minipage}{0.42\textwidth}
\begin{center}
            \begin{minipage}{0.20\textwidth}
              \centerline{\includegraphics[width=\textwidth]{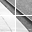}}
            \end{minipage}

            \begin{minipage}{1\textwidth}
              \centerline{\includegraphics[width=\textwidth]{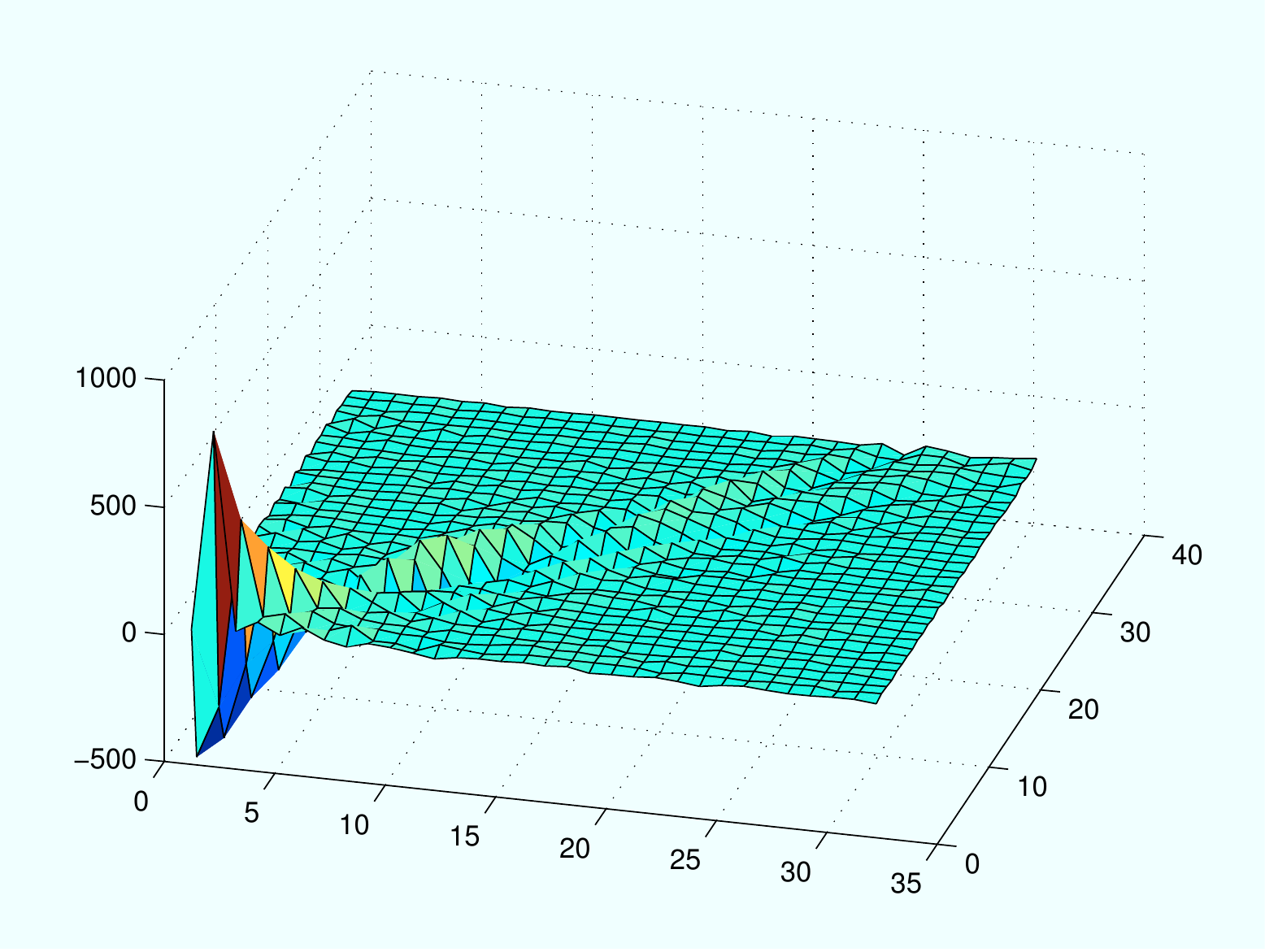}}
            \end{minipage}
\end{center}
        \end{minipage}
     
      \end{center}
    \end{minipage}

  \begin{minipage}{1.0\textwidth}
    \begin{minipage}{0.1\textwidth}
      \centerline{\bf(i)}
    \end{minipage}
    \hfill
    \begin{minipage}{0.42\textwidth}
      \centerline{\bf(ii)}
    \end{minipage}
    \hfill
    \begin{minipage}{0.42\textwidth}
      \centerline{\bf(iii)}
    \end{minipage}
  \end{minipage}

~
    \caption
      {
      An example of the processing done in Section~\ref{subsec:Calculation_of_energy_potential}.
      {\bf (i)}~A clique involving empty node $X$ with two candidate labels in its representative set.
      {\bf (ii)}~A clique and a graphical representation of its DCT coefficient matrix
      where node $X$ is initialised with candidate label~1.
      The gaps between the blocks are for ease of interpretation only
      and are not present during DCT calculation. 
      {\bf(iii)}~As per~(ii), but using candidate label~2. 
      The smoother spectral distribution for candidate~2 suggests
      that it is a better fit than candidate~1.
      }
    \label{fig:FreqDist}
\end{figure*}

Similarly, for other labels present in the state space of node {$X$}, we compute their corresponding 2D DCT as mentioned above.
A graphical example of the procedure is shown in Fig.~\ref{fig:FreqDist}.

Assuming that pixels close together have similar intensities, 
when the correct label is placed at node {$X$}, 
the resulting transformation has a smooth response (less high frequency components) when compared to other 
candidate labels.

The energy potential for each label is calculated after summing potentials
obtained across the $\phi$ colour channels, as given below:
\begin{equation}
{
  V_c(\omega_k) =
 \sum\nolimits_{z = 1}^{\phi}
  \left(
    \sum\nolimits_{v = 1}^{M} \sum\nolimits_{u = 1}^{M} \left| T_k^{z}{(v,u)} \right|
  \right)}
\label{eqn:energypotential}
\end{equation}

\noindent
where $\omega_k$ is the local configuration involving label $k$.
The potentials over the other three cliques in Fig.~\ref{fig:clique1}
are calculated in a similar manner.

\subsection{Modified Background Model for Foreground Segmentation}
\label{subsec:Modified_Background_Model}

The foreground detection framework described in Section~\ref{sec:Background_Subtraction_Framework}
uses a background model comprised of location specific multivariate Gaussians. 
The background image reconstructed through the MRF-based process is used as follows.
First, the dual-Gaussian training strategy used in Section~\ref{sec:Background_Subtraction_Framework} 
is run on a given training sequence, obtaining the mean vectors and diagonal covariance matrices for each location.
The mean vectors are then replaced by rerunning step 1 of the segmentation framework
on the estimated background image.
The covariance matrices are retained as is.
Preliminary experiments indicated that when stationary backgrounds were occluded
by foreground objects for a long duration,
the variances computed in step 1 were similar to the variances of the true background.


\section{Experiments}
\label{sec:Experiment_Results}

The proposed framework (MRF {\footnotesize{+}} multi-stage classifier)
was evaluated with segmentation methods based on the baseline multi-stage classifier~\cite{Reddy_TCSVT_inpress},
Gaussian mixture models (GMMs)~\cite{stauffer1999abm} and 
feature histograms~\cite{li2003foreground}.
In our experiments the same parameter settings were used across all sequences
(i.e.,~they were not optimised for any particular sequence). The block size was set to 16 $\times$ 16.
The values of $\mathcal{T}_1$ and $\mathcal{T}_2$ (see Eqns.~\ref{eqn:corr_ptrk}~and~\ref{eqn:diff_ptrk})  
were set to 0.8 and 3 respectively,
while $W_{max}$~(see Eqn.~\ref{eqn:rayleigh}) and $T$ (Eqn.~\ref{eqn:mrf_defines1})
were set to 150 and 1024 respectively.
The algorithm was implemented in C++ with the aid of the Armadillo library~\cite{Armadillo}.

We used the OpenCV~v2.0~\cite{Bradski2008} implementations for the last two algorithms,
in conjunction with morphological post-processing
(opening followed by closing using a {\small $3\hspace{-1pt}\times\hspace{-1pt}3$} kernel)
in order to improve the quality of the obtained foreground masks~\cite{li2003foreground}.
The methods' default parameters were found to be optimal,
except for the histogram method, where the built-in morphology operation was disabled
as we found that it produced worse results than the above-mentioned opening and closing.
We note that the proposed foreground segmentation approach does not require any such ad hoc post-processing.

In our experiments, we studied the influence of the various foreground segmentation algorithms on tracking performance.
The foreground masks obtained from the detectors were passed as input to several tracking systems.
We used the tracking systems implemented in the video surveillance module of OpenCV v2.0~\cite{Bradski2008}
and the tracking ground truth data that is available for the sequences
in the second set of the {\text{\small{CAVIAR}}}%
\footnote{http://homepages.inf.ed.ac.uk\hspace{1pt}/\hspace{1pt}rbf\hspace{1pt}/\hspace{1pt}CAVIARDATA1\hspace{1pt}/}
dataset.
We randomly picked 30 sequences from the dataset for our experiments.
The tracking performance was measured with two metrics:
multiple object tracking accuracy (MOTA) and multiple object tracking precision (MOTP),
as proposed by Bernardin and Stiefelhagen~\cite{BernardinEtAl2008}.

\begin{figure*}[!tb]
\begin{minipage}{1.0\textwidth}

  \begin{minipage}{1.0\textwidth}
  
      \includegraphics[width=0.1935\columnwidth]{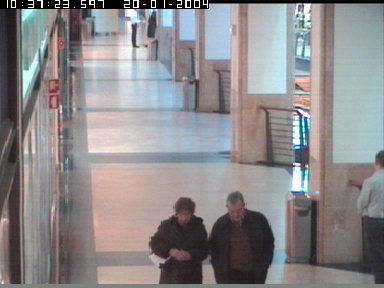}
      \hfill
      \includegraphics[width=0.1935\columnwidth]{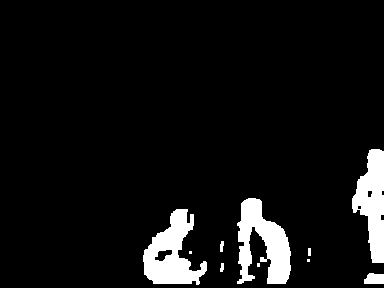}
      \hfill
      \includegraphics[width=0.1935\columnwidth]{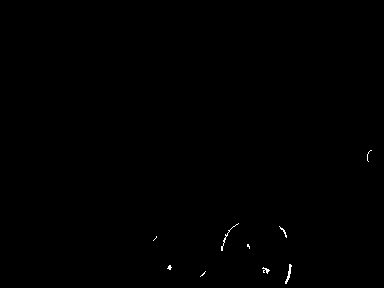}
      \hfill
      \includegraphics[width=0.1935\columnwidth]{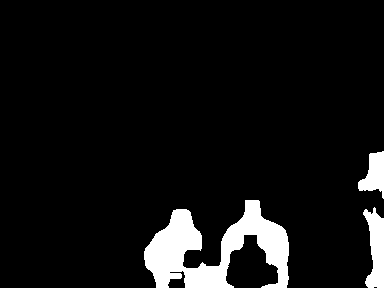}
      \hfill      
      \includegraphics[width=0.1935\columnwidth]{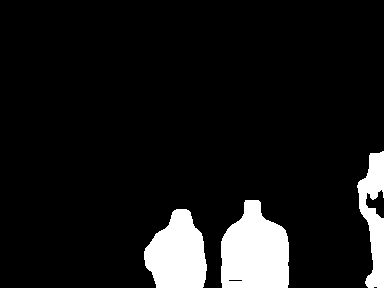}
    
  \end{minipage}

    ~       

  \begin{minipage}{1.0\textwidth}
  
      \includegraphics[width=0.1935\columnwidth]{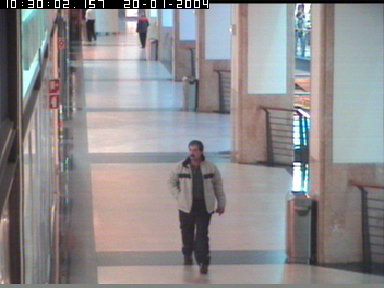}
      \hfill
      \includegraphics[width=0.1935\columnwidth]{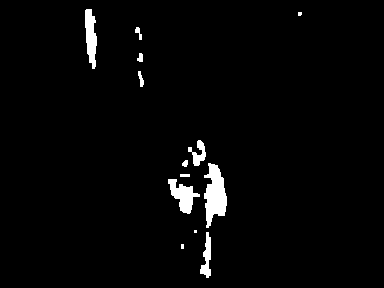}
      \hfill
      \includegraphics[width=0.1935\columnwidth]{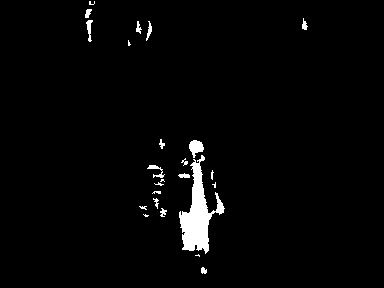}
      \hfill
      \includegraphics[width=0.1935\columnwidth]{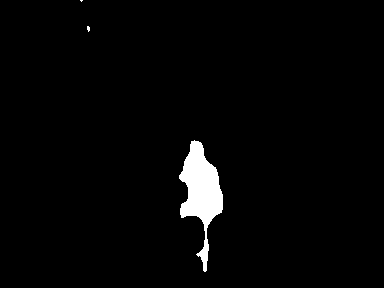}
      \hfill      
      \includegraphics[width=0.1935\columnwidth]{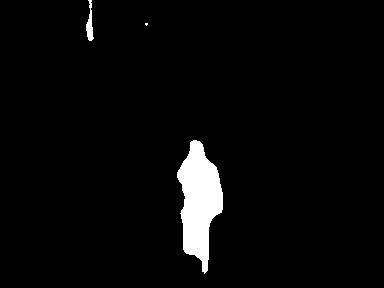}
    
  \end{minipage}

    ~     
    
  \begin{minipage}{1.0\textwidth}
  
      \includegraphics[width=0.1935\columnwidth]{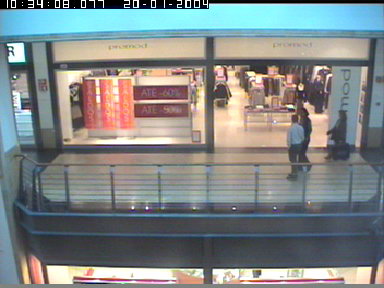}
      \hfill
      \includegraphics[width=0.1935\columnwidth]{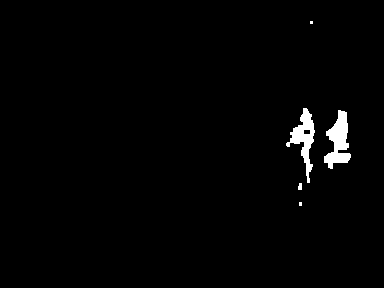}
      \hfill
      \includegraphics[width=0.1935\columnwidth]{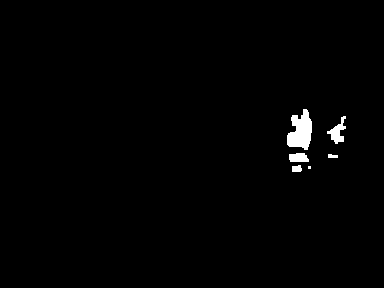}
      \hfill
      \includegraphics[width=0.1935\columnwidth]{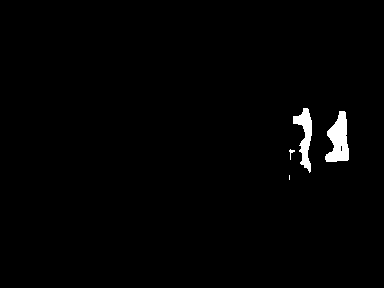}
      \hfill      
      \includegraphics[width=0.1935\columnwidth]{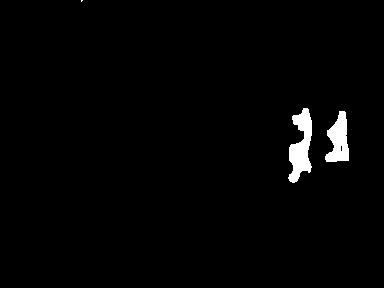}
    
  \end{minipage}

~

  \begin{minipage}{1.0\textwidth}
    \begin{minipage}{0.1935\columnwidth}
      \centerline{\bf(i)}
    \end{minipage}
    \hfill
    \begin{minipage}{0.1935\columnwidth}
      \centerline{\bf(ii)}
    \end{minipage}
    \hfill
    \begin{minipage}{0.1935\columnwidth}
      \centerline{\bf(iii)}
    \end{minipage}
    \hfill
    \begin{minipage}{0.1935\columnwidth}
      \centerline{\bf(iv)}
    \end{minipage}
    \hfill
    \begin{minipage}{0.1935\columnwidth}
      \centerline{\bf(v)}
    \end{minipage}
    
  \end{minipage}

~
  
  \caption
    {
    {\bf (i)}~Example frames from CAVIAR dataset;
    foreground masks obtained using: {\bf (ii)} GMM based method~\cite{stauffer1999abm},
    {\bf (iii)} histogram based method~\cite{li2003foreground},
    {\bf (iv)} baseline multi-stage classifier~\cite{Reddy_TCSVT_inpress},
    {\bf (v)}~ proposed MRF based framework.
    We note the masks shown in columns {\bf (ii)} to {\bf (iv)} have considerable amount of 
    false negatives since the foreground objects were included in the background model,
    while the results of the proposed framework (column {\bf (v)}) have minimal errors.
    }
  \label{fig:fg_detection_with_3_ea}
     \end{minipage} 

~

~

\end{figure*}
 
Briefly, 
MOTP measures the average pixel distance between the ground-truth locations of objects
and their locations according to a tracking algorithm.
The lower the MOTP, the better.
MOTA accounts for object configuration errors, false positives, misses as well as mismatches.
The higher the MOTA, the better.

We performed 20 tracking simulations by evaluating four foreground object segmentation algorithms 
(baseline multi-stage classifier, GMM, feature histogram and the proposed method)
in combination with five tracking algorithms
(blob matching, mean shift, mean shift with foreground feedback, particle filter,
and blob matching with particle filter for occlusion handling).
The performance result in each simulation is the average performance of the 30 test sequences.
We used the first 200 frames of each sequence for initialising the background model.

Examples of qualitative results are illustrated in~Fig.~\ref{fig:fg_detection_with_3_ea}.
It can be observed that foreground masks generated using methods based on
GMMs~\cite{stauffer1999abm}, 
feature histograms~\cite{li2003foreground},
and the baseline multi-stage classifier~\cite{Reddy_TCSVT_inpress}
have considerable false negatives,
which are due to foreground objects being included into the background model.
In contrast, the MRF based model initialisation approach results in noticeably better foreground detection. 

\begin{figure*}[!tb]
  \centering
  \begin{minipage}{1\textwidth}
    \begin{minipage}{0.03\textwidth}
      \begin{flushright}
        {\footnotesize (a)}
      \end{flushright}
    \end{minipage}
    \hfill
    \begin{minipage}{0.97\textwidth}
      \centerline{\includegraphics[height=0.28\columnwidth,width=\columnwidth]{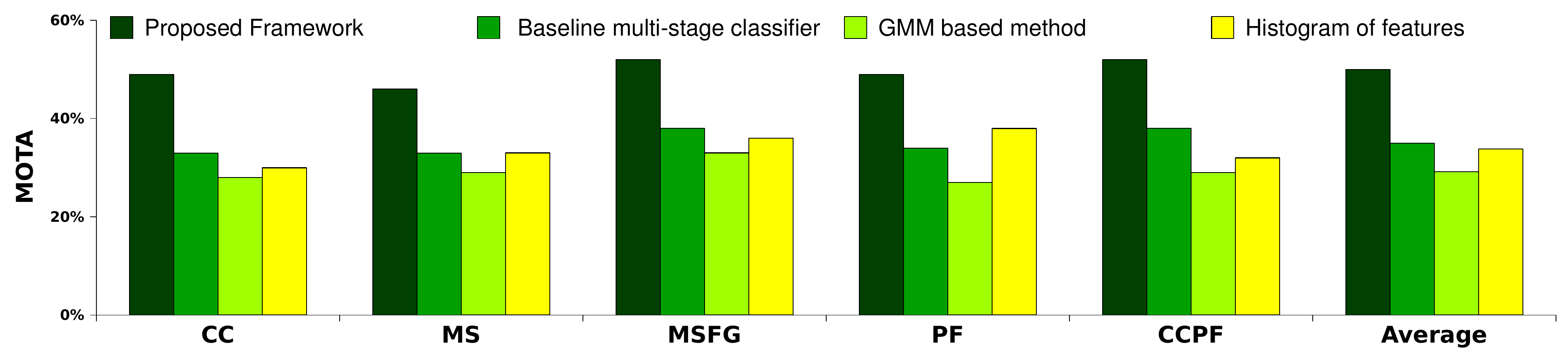}}
    \end{minipage}
  \end{minipage}
  \begin{minipage}{\textwidth}
    \begin{minipage}{0.03\textwidth}
      \begin{flushright}
        {\footnotesize (b)}
      \end{flushright}
    \end{minipage}
    \hfill
    \begin{minipage}{0.97\textwidth}
      \centerline{\includegraphics[height=0.29\columnwidth,width=\columnwidth]{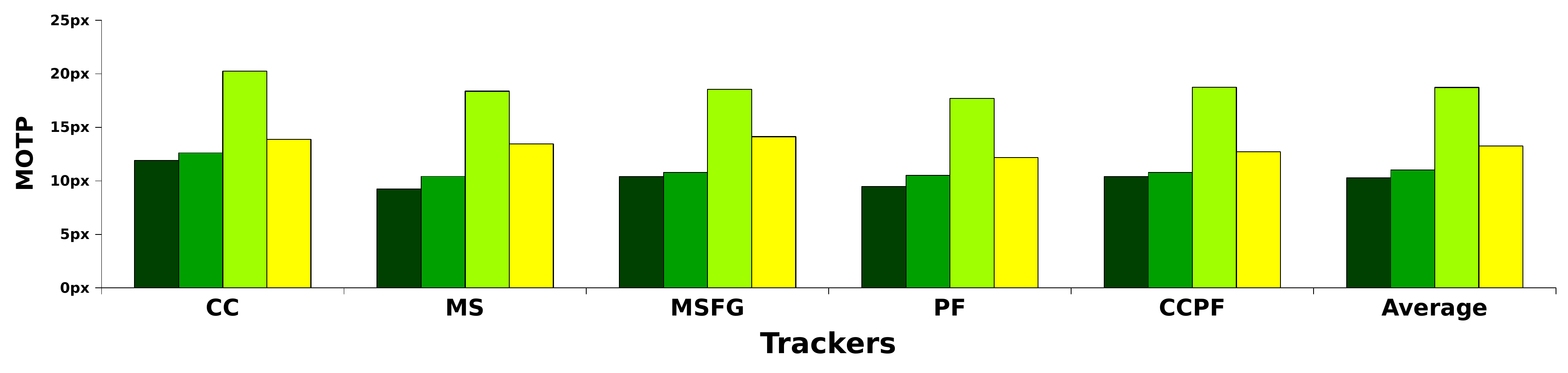}}
    \end{minipage}
  \end{minipage}
  \caption
    {
    \small
    Effect of foreground detection methods on: 
    {\bf (a)} multiple object tracking accuracy (MOTA), where taller bars indicate better accuracy;
    {\bf (b)} multiple object tracking precision (MOTP), where shorter bars indicate better precision (lower distance).
    Results are grouped by tracking algorithm:
    blob matching~(CC),
    \mbox{mean shift trackers~(MS and MSFG)},
    particle filter~(PF)
    and hybrid tracking~(CCPF).
    }
  \label{fig:mota_motp}
\end{figure*}

The quantitative tracking results, presented in Fig.~\ref{fig:mota_motp},
indicate that in all cases the proposed framework
led to the best precision and accuracy values.
For tracking precision (MOTP),
the next best method~\cite{Reddy_TCSVT_inpress} 
obtained an average pixel distance of 11.03,
while the proposed method reduced the distance to~10.28,
indicating an improvement of approximately 7\%.
For tracking accuracy (MOTA),
the next best method 
obtained an average accuracy value of 0.35,
while the proposed method achieved 0.5, 
representing a considerable improvement of about~43\%. 

\newpage
\section{Main Findings}
\label{sec:Conclusion}

In this paper we have proposed a foreground segmentation framework which effectively 
segments foreground objects in cluttered environments.
The MRF-based model initialisation strategy allows the training sequence to contain foreground objects.
We have shown that good background model initialisation results 
in considerably improved foreground detection, which leads to better tracking.

We noticed (via subjective observations) that all evaluated algorithms perform reasonably well
when foreground objects are always in motion
(i.e., where the background is visible for a longer duration when compared to the foreground).
However, accurate estimation by methods solely relying on temporal statistics to initialise
their background model becomes problematic if the above condition is not satisfied.
This is the main area where the proposed framework is able to detect foreground objects accurately.

A minor limitation exists, as there is a potential to mis-estimate the background
in cases where an occluding foreground object is smooth (uniform intensity value),
has intensity value similar to that of the background
(i.e., low contrast between the foreground and the background)
and the true background is characterised by strong edges.
Under these conditions, the energy potential of the label 
containing the foreground object is smaller (i.e., smoother spectral response)
than that of the label corresponding to the true background.
This limitation will be addressed in future work.

Overall, the parameter settings for the proposed algorithm appear to be quite robust against a variety of sequences
and the method does not require explicit post-processing of the foreground masks.
Experiments conducted to evaluate the effect on tracking performance (using the {\text{\small{CAVIAR}}} dataset)
show the proposed framework obtains considerably better results (both qualitatively and quantitatively)
than approaches based on
Gaussian mixture models (GMMs)~\cite{stauffer1999abm} and 
feature histograms~\cite{li2003foreground}.


\section*{Acknowledgements}

\begin{small}
NICTA is funded by the Australian Government as represented by the {\it Department of Broadband, Communications and the Digital Economy}
as well as the Australian Research Council through the {\it ICT Centre of Excellence} program.
\end{small}


\balance
\begin{small}
\bibliographystyle{ieee}
\bibliography{references}
\end{small}


\end{document}